\documentclass[conference]{IEEEtran}
\IEEEoverridecommandlockouts

\usepackage{amsmath}
\usepackage{algorithm}
\usepackage{algpseudocode}
\usepackage{graphicx, subfig}
\usepackage{booktabs}
\usepackage{multirow}
\usepackage{array}
\usepackage{url}
\usepackage{color}

\newcommand{\tabincell}[2]{\begin{tabular}{@{}#1@{}}#2\end{tabular}}

\def\BibTeX{{\rm B\kern-.05em{\sc i\kern-.025em b}\kern-.08em
    T\kern-.1667em\lower.7ex\hbox{E}\kern-.125emX}}
\begin{document}

\title{Protecting the Intellectual Properties of Deep Neural Networks with an Additional Class and Steganographic Images}

\author{\IEEEauthorblockN{Shichang Sun\textsuperscript{1},
        Mingfu Xue\textsuperscript{1},
        Jian Wang\textsuperscript{1},
        and Weiqiang Liu\textsuperscript{2}
      }
\IEEEauthorblockA{\textsuperscript{1}College of Computer Science and Technology, Nanjing University of Aeronautics and Astronautics, Nanjing, China}
\IEEEauthorblockA{\textsuperscript{2}College of Electronic and Information Engineering, Nanjing University of Aeronautics and Astronautics, Nanjing, China}
\{sunshichang, mingfu.xue, wangjian, liuweiqiang\}@nuaa.edu.cn
}

\maketitle

\begin{abstract}
Recently, the research on protecting the intellectual properties (IP) of deep neural networks (DNN) has attracted serious concerns. A number of DNN copyright protection methods have been proposed, including embedding watermarks into DNN's parameters, constructing watermarking based on backdoor, and so on.
However, most of the existing watermarking methods focus on verifying the copyright of the model, which do not support the authentication and management of users' fingerprints, thus
can not satisfy the requirements of commercial copyright protection.
In addition, the query modification attack which was proposed recently can invalidate most of the existing backdoor-based watermarking methods.
To address these challenges, in this paper, we propose a method to protect the intellectual properties of DNN models by using an additional class and steganographic images.
Specifically, we use a set of watermark key samples to embed an additional class into the DNN, so that the watermarked DNN will classify the watermark key sample as the predefined additional class in the copyright verification stage.
We adopt the least significant bit (LSB) image steganography to embed users' fingerprints into watermark key images.
Each user will be assigned with a unique fingerprint image so that the user's identity can be authenticated later.
Experimental results demonstrate that, the proposed method can protect the copyright of DNN models effectively.
On Fashion-MNIST and CIFAR-10 datasets, the proposed method can obtain 100\% watermark accuracy and 100\% fingerprint authentication success rate.
In addition, the proposed method is demonstrated to be robust to the model fine-tuning attack, model pruning attack, and the query modification attack.
Compared with three existing watermarking methods (the logo-based, noise-based, and adversarial frontier stitching watermarking methods), the proposed method has better performance on watermark accuracy and robustness against the query modification attack.
\end{abstract}


\section{Introduction}
Deep neural networks (DNN) have achieved significant success in computer vision tasks such as image classification, object detection, and face recognition.
Furthermore, machine learning as a service (MLaaS) \cite{RibeiroGC15} provided by large companies, has also become a popular business paradigm.
However, illegal users can pirate and abuse models, such as stealing pre-trained deep learning models, and build pirated AI services \cite{ZhangGJWSHM18}.
In this way, the intellectual property (IP) of the model owner will be infringed.
Protecting the intellectual properties of deep neural networks is a challenging task.
The adversaries can take some measures to prevent the pirated models from being discovered by the model owner.
For example, an adversary can tamper with the pirated DNN model with model fine-tuning \cite{Pittaras2017} or model pruning \cite{HanPTD15}, thereby removing the potential watermark.

Existing copyright protection methods \cite{ZhangGJWSHM18, UchidaNSS17, MerrerPT20, AdiBCPK18} use watermarking technique to protect the ownership of the model owner.
The watermarks can be embedded in the parameters of the DNN, or constructed based on backdoor, and so on.
For example, Uchida \textit{et al.} \cite{UchidaNSS17} embed the watermark into the parameters of the deep neural network, but this method can only be applied in the white-box scenarios.
Adi \textit{et al.} \cite{AdiBCPK18} proposed a DNN watermarking method based on the backdoor.
Zhang \textit{et al.} \cite{ZhangGJWSHM18} proposed three backdoor based DNN watermarking methods.
Merrer \textit{et al.} \cite{MerrerPT20} proposed an adversarial example based DNN watermarking method.
These watermarking methods \cite{ZhangGJWSHM18, MerrerPT20, AdiBCPK18} can be used to perform ownership verification remotely in the black-box scenarios.
However, most of the existing watermarking methods \cite{ZhangGJWSHM18, UchidaNSS17, MerrerPT20, AdiBCPK18} do not support the user's identity authentication, thus are not suitable for practical commercial applications.
In addition, a recent work \cite{NambaS19} has shown that, watermarking methods such as logo-based watermarking method and noise-based watermarking method are vulnerable to the query modification attack.
In the query modification attack, the attacker employs an autoencoder to detect and modify the watermarked images, thereby invalidating the watermark verification process of the model owner \cite{NambaS19}.
The autoencoder is a special deep neural network, which first compresses the input images into the low-dimensional data, and then decompresses these low-dimensional data as the output images \cite{NambaS19}.
As a result, the noise in input images can be removed.
In this way, the logo or noise (i.e., watermark patterns) in watermark key samples can be removed by the query modification attack.

In this paper, we propose to embed the watermark in the DNN by using an additional class.
We select a small number of clean images outside the original training dataset as the watermark key samples.
Subsequently, a user's fingerprint will be hidden in each watermark key sample.
To embed the watermark, we first assign a new class label to all the watermark key samples, and then add the watermark key samples to the training set to train a watermarked DNN.
The trained DNN (i.e., the watermarked DNN) can output an extra class when a watermark key sample is input.
Since we use clean images outside the training set as the watermark key samples rather than superimposing patterns on the images, the proposed watermarking method can resist the query modification attack \cite{NambaS19}.
In order to support user's fingerprint authentication, we use image steganography \cite{lsbsteganography2020} to hide users' fingerprints in watermark key samples.
Specifically, we use the least significant bit (LSB) \cite{lsbsteganography2020} method to embed different users' fingerprints into different watermark key samples, and distribute a unique fingerprint image to each authorized user.

The contributions of this paper are as follows:
\begin{itemize}
\item
\textbf{Adding an additional class to protect the copyright of DNN.}
We propose to embed the watermark into the DNN model by using an extra class.
To verify the ownership, the model owner can input the watermark key sample to the DNN to trigger the additional class.
Experimental results show that, the proposed method can achieve 100\% watermark accuracy \cite{ZhangGJWSHM18} on two Fashion-MNIST \cite{abs-1708-07747} and CIFAR-10 \cite{krizhevsky2009learning} datasets without affecting the test accuracy of the model.
Compared with the existing watermarking methods \cite{ZhangGJWSHM18, MerrerPT20}, the proposed watermarking method can obtain a higher watermark accuracy and is more robust to the query modification attack \cite{NambaS19}.

\item
\textbf{Supporting user's identity authentication.}
We hide the user's fingerprint in the image by using the LSB \cite{lsbsteganography2020} image steganography method. Users' fingerprints can be extracted to verify their identities for access control.
Experimental results show that the fingerprint authentication success rate (FASR) of the proposed method is 100\%.
Meanwhile, the user's fingerprint embedded in the image will not affect the image quality, as the mean squared error (MSE) \cite{WangBSS04} of steganographic images is only 0.044 (Fashion-MNIST) and 0.015 (CIFAR-10), respectively.
\item
\textbf{Robust to query modification attack.}
In addition to be robust to model fine-tuning and model pruning, the proposed watermarking method can also resist the query modification attack \cite{NambaS19}.
Under the query modification attack, the proposed watermarking method can still obtain 100\% watermark accuracy on the LeNet-5 \cite{726791} model and 85\% watermark accuracy on the VGG-16 \cite{SimonyanZ14a} model, which are significantly higher than the related works.
As a comparison, the logo-based watermarking method \cite{ZhangGJWSHM18} only obtains 11\% (LeNet-5) and 15\% (VGG-16) watermark accuracy, the noise-based method \cite{ZhangGJWSHM18} only obtains 14\% (LeNet-5) and 16\% (VGG-16) watermark accuracy, and the adversarial frontier stitching watermarking method \cite{MerrerPT20} obtains 91\% (LeNet-5) and 35\% (VGG-16) watermark accuracy.
\end{itemize}

\section{Related Work}
In recent years, many watermarking works have been proposed in the context of DNN copyright protection.
We briefly review these existing watermarking methods as follows.

Uchida \textit{et al.} \cite{UchidaNSS17} developed a watermark embedding and extraction method based on parameter regularizer. Under the adjustment of the parameter regularizer, the watermark was embedded in the weight of the DNN without affecting the performance of the DNN.
Rouhani \textit{et al.} \cite{RouhaniCK19} embedded the watermark into the probability density function of activation maps of the DNN model.

Merrer \textit{et al.} \cite{MerrerPT20} proposed a watermarking method based on adversarial examples.
Specifically, they crafted some adversarial examples, and then retrained the DNN with these adversarial examples.
During ownership verification, if the number of correctly classified adversarial examples is greater than the predefined threshold, the copyright of the model can be verified \cite{MerrerPT20}.
Adi \textit{et al.} \cite{AdiBCPK18} proposed to protect the IP of DNN models by using backdoor.
In this method, the watermarked DNN is trained to output incorrect label for particular abstract images.
The model owner can send these abstract images to the remote DNN model to verify whether the model carries a watermark \cite{AdiBCPK18}.
Similarly, Zhang \textit{et al.} \cite{ZhangGJWSHM18} proposed to embed three different types of watermarks into the DNN model through the backdoor technique.
The three types of watermarks are the logo-based watermark, the noise-based watermark, and the unrelated watermark, respectively.
Zhong \textit{et al.} \cite{ZhongZ0G020} proposed to embed the watermark (the logo pattern) into the DNN with the help of a new class label.

This paper aims to watermark the DNN with an additional class and steganographic images. Compared with the existing watermarking methods, our method has the following advantages:
1) support the user's fingerprint authentication;
2) be robust to the state-of-the-art query modification attack;
3) can achieve better performance on watermark accuracy \cite{ZhangGJWSHM18} than the works \cite{ZhangGJWSHM18, MerrerPT20}.

\section{The Proposed Method} \label{the_proposed_method}
In this section, we propose a method to protect the intellectual property of DNN models by using an additional class and steganographic images.
We elaborate the proposed method from four aspects: watermark key sample generation, watermark embedding, ownership verification, and user's fingerprint authentication.

\subsection{Overview} \label{overview}
\begin{figure}[!t]
\centering
    \includegraphics[width=3.5in]{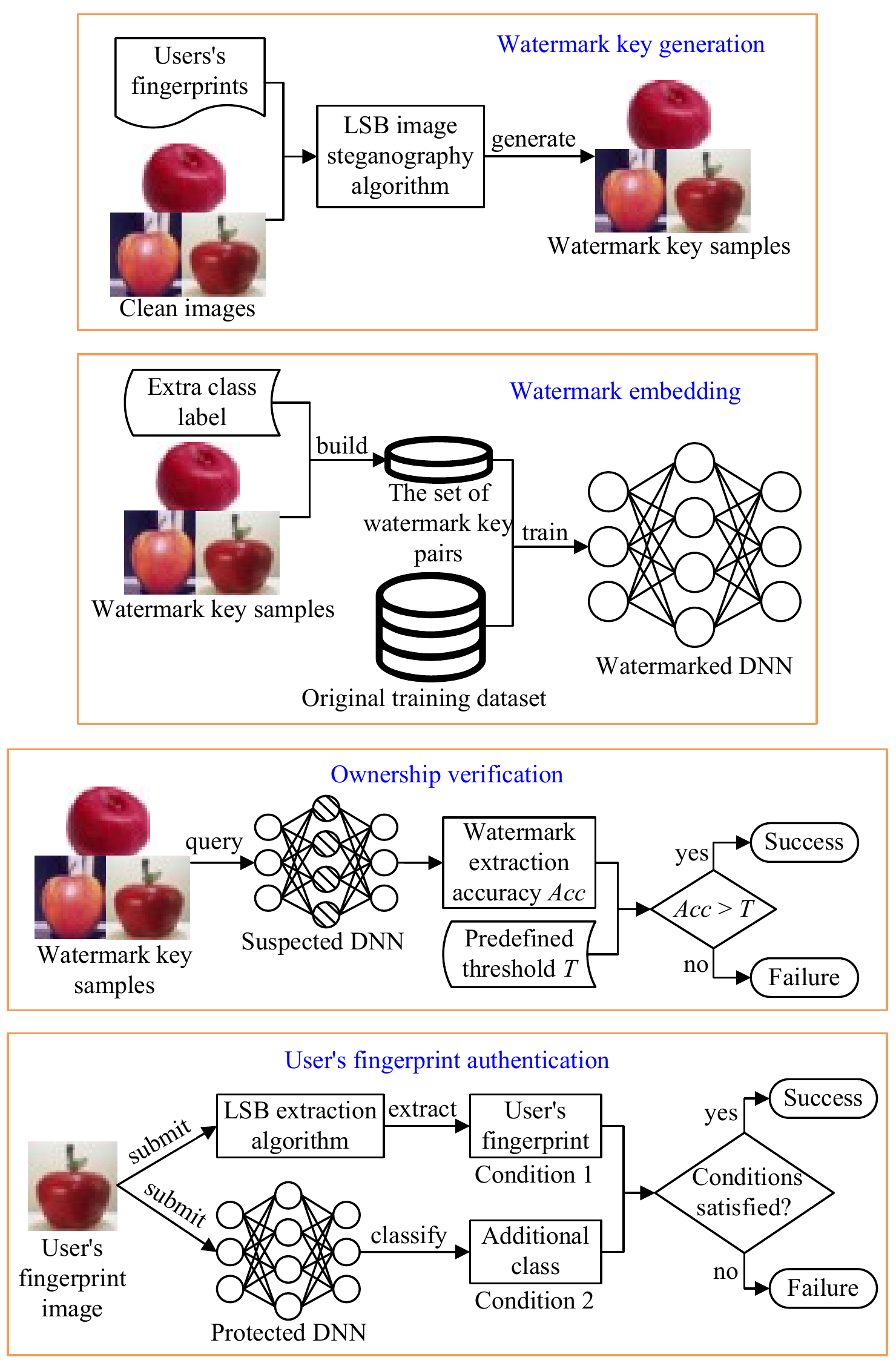}
\caption{Overall flow of the proposed method.}
\label{fig1}
\end{figure}

The overall flow of the proposed watermarking method is shown in Fig. \ref{fig1}.
It can be divided into four stages: watermark key sample generation, watermark embedding, ownership verification, and user's fingerprint authentication.
To protect DNN's intellectual property,
first, the watermark key samples are generated from some meaningful images (e.g., the apple images in Fig. \ref{fig1}) outside the original training set.
Additionally, a user's fingerprint is hidden in a watermark key sample via the LSB steganography \cite{lsbsteganography2020}.
Second, in the training phase,
the generated watermark key samples assigned with an extra class label are injected into the training set to embed the watermark into the DNN model.
Third, the model owner can remotely verify the ownership by sending watermark key samples to the suspected model.
Lastly, the watermark key sample can serve as a fingerprint image to verify the user's identity.

\subsection{Watermark Key Sample Generation} \label{watermark_key_generation}
In this section, we introduce how to generate watermark key samples.
First, a small amount of clean images outside the original training set are randomly selected as the watermark key samples.
Second, different users' fingerprints are embedded in the watermark key images with the image steganography, where each watermark key sample is embedded with a user's fingerprint.

The user's fingerprint is composed of a binary string, which can uniquely represent the user's identity.
For example, in the experiments, the character string ``user fp\textit{i}'' is used as the fingerprint of the \textit{i}-th user.

We adopt the least significant bit (LSB) steganography \cite{lsbsteganography2020} to embed users' fingerprints into watermark key images.
In order to hide a message inside the image, the LSB image steganography method will replace the least significant bit (i.e., the last bit) of pixel values with the bits of the message.

\begin{figure}[!htbp]
\centering
\includegraphics[width=3.5in]{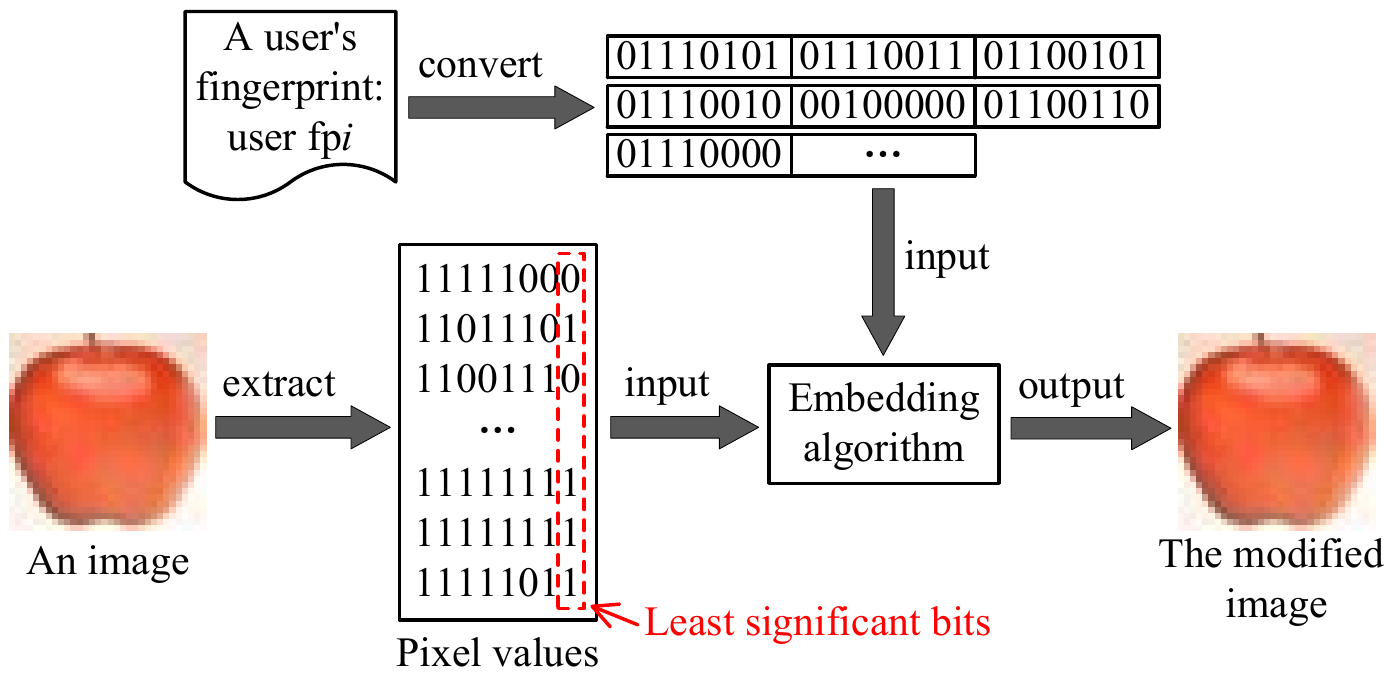}
\caption{The process of embedding a user's fingerprint by using the LSB steganography.}
\label{fig2}
\end{figure}

The process of embedding the user's fingerprint into an image by using the LSB algorithm is shown in Fig. \ref{fig2}.
For a user's fingerprint, each character of the fingerprint will be converted into binary bits.
For an image, it contains $w \times h$ pixels, and each pixel value consists of 8 bits (0$\sim$255), where $w$ and $h$ are the width and height of the image in pixels.
The LSB algorithm hides the user's fingerprint by replacing the lowest bit with each bit of the fingerprint.
Moreover, if the least significant bits of all pixel values are filled up, the LSB algorithm will use the next least significant bit (i.e., the second-lowest bit) of each pixel to embed the remaining fingerprint information.
Since an image has plenty of pixels ($w \times h$ pixels), the LSB steganography is effective for embedding enough fingerprint information.

\subsection{Watermark Embedding} \label{watermark_embedding}
After generating the watermark key samples, we embed an additional class (e.g., the ``apple'' class) into the DNN as the watermark.
The additional class is added to the DNN, which is not one of the classes in DNN's main classification tasks.

The watermark embedding algorithm is shown in Algorithm \ref{alg1}.
It takes the original training set $D$, a watermark key set $W$, an additional class label $\hat y$, and the users' fingerprint set ${FP}$ as input, and then outputs the watermarked DNN model $M_{wm}$.
To embed the watermark, first, $K$ users' fingerprints are injected into $K$ samples of $W$  by using the LSB algorithm, respectively (Line 1$\sim$3 in Algorithm \ref{alg1}).
Second, the additional class label $\hat y$ is assigned to each data in $W$ to generate the set of watermark key pairs $D_{wm}$ (Line 4$\sim$5 in Algorithm \ref{alg1}).
Third, the model owner adds the set $D_{wm}$ to $D$ to build the training set $D_{train}$, and then trains the DNN model (Line 7$\sim$8 in Algorithm 1).
After the training process, the watermark will be embedded into the DNN, and accordingly, the watermarked DNN can output the extra class when the specific input (i.e., the watermark key samples) arrives in the inference phase.
Note that, in our method, the DNN model learns the pattern of the watermark from the image features instead of the users' fingerprints.
In other words, the watermarked DNN classifies a watermark key image into the additional class instead of classifying a user's fingerprint information (``user fp\textit{i}'')  into the additional class.

\begin{algorithm}[!htbp]
\caption{Watermark embedding algorithm}
\label{alg1}
\renewcommand{\algorithmicrequire}{\textbf{Input:}}
\renewcommand{\algorithmicensure}{\textbf{Output:}}
\begin{algorithmic}[1]
\Require{Original training set $D = \{ {x_i},{y_i}\} _{i = 1}^N$, watermark key set $W = \{ w{_i}\} _{i = 1}^K$, additional class label $\hat y$, users' fingerprint set $FP = \{ {f_i}\} _{i = 1}^K$, the set of watermark key pairs $D_{wm}$}
\Ensure{Watermarked DNN model $M_{wm}$}
\State \textbf{initialize}: ${D_{wm}} = \emptyset$;
\For {$i=1, 2, \dots, K$}
    \State ${w_i} = LSB(w{_i},{f_i})$;
    \State $y_i = \hat y$;
    \State $D_{wm} = {D_{wm}} \cup \{ {w_i},{y_i}\}$;
\EndFor
\State $D_{train} = D \cup D_{wm}$;
\State $M_{wm} = Train(D_{train})$;
\State \Return $M_{wm}$
\end{algorithmic}
\end{algorithm}

\subsection{Ownership Verification} \label{ownership_verification}
When the watermarked DNN model is pirated by an adversary, the model owner can query the pirated DNN model remotely through prediction application programming interfaces (APIs).
Specifically, the model owner sends the watermark key samples to the suspected model $M_{sp}$, and obtains the predicted classes that returned from $M_{sp}$.
If the accuracy of $M_{sp}$ on the set $D_{wm}$ is greater than the predefined threshold $T$, the model owner can determine that the model $M_{sp}$ is a pirated model. Otherwise, $M_{sp}$ is not a pirated model.
Formally, the model owner can successfully verify the ownership of $M_{sp}$ if the following condition is satisfied:
\begin{equation}
Acc(M_{sp}, D_{wm}) > T
\label{equ1}
\end{equation}
where $Acc$ denotes the model's (i.e., $M_{sp}$) accuracy on the set of watermark key pairs $D_{wm}$, and $T$ is the predefined threshold for ownership verification.

The reasons why the proposed watermarking method can works for ownership verification are as follows.
i) Reliability: the watermark in the DNN can only be triggered by watermark key samples that are private to the model owner.
ii) Robustness: even if the DNN model is pirated or slightly modified, the watermark can still be retained in the model.
iii) Integrity: for a clean DNN model, since it does not contain the watermarked class (the additional class), it will not return the correct result when a watermark key sample is input.

\subsection{User's Fingerprint Authentication} \label{user_fp_authentication}
To support the multi-users' identity authentication, each authorized user is assigned with a watermark key sample (i.e., an fingerprint image) as the fingerprint.
A user is granted legal access to the DNN model only after his or her identity is verified.
In other words, before using the DNN model, the user is required to submit the fingerprint image for fingerprint authentication.
The effectiveness of user's fingerprint authentication should satisfy the following two conditions simultaneously:
1) the user's fingerprint can be extracted from the image submitted by the user;
2) the submitted fingerprint image is classified as the additional class by the watermarked model.
The above two conditions can be formalized as:
\begin{equation}
\left \{ {\begin{array}{l}
  {LSB\_Extraction(w_f) \in FP}\vspace{0.6ex} \hfill \\
  {{M_{wm}}(w_f) = \hat y} \hfill
\end{array}} \right.
\label{equ2}
\end{equation}
where $w_f$ is the fingerprint image submitted by the user, and $LSB\_Extraction$ denotes the message extraction function of the LSB steganography \cite{lsbsteganography2020}, which is the reverse process of the LSB embedding algorithm.

In our method, only when the above two conditions are satisfied simultaneously, the user can successfully pass the fingerprint authentication.
It is difficult for an attacker without the knowledge of the proposed method to forge a user's fingerprint.
Additionally, since each fingerprint image is different from the other, it is difficult for multiple users to cooperate with each other to derive the steganographic information in multiple fingerprint images.

\section{Experiments}\label{experiments}
In this section, we evaluate the performance of the proposed method.
First, we introduce the experimental settings, including datasets and models, parameter settings, and evaluation metrics.
Then, we analyze the effectiveness of the proposed method from two aspects: ownership verification and user's fingerprint authentication.
Next, we demonstrate the robustness of the proposed method against model fine-tuning attack \cite{Pittaras2017}, model pruning attack \cite{HanPTD15}, and query modification attack \cite{NambaS19}.
Finally, we compare the proposed method with three watermarking methods, which are logo-based \cite{ZhangGJWSHM18}, noise-based \cite{ZhangGJWSHM18}, and adversarial frontier stitching (AFS) \cite{MerrerPT20} watermarking methods.

\subsection{Experimental Setup} \label{experimental_setup}
\textbf{Datasets and models}.
We evaluate the proposed watermarking method on two datasets (Fashion-MNIST \cite{abs-1708-07747} and CIFAR-10 \cite{krizhevsky2009learning}).
For the Fashion-MNIST dataset, we train the LeNet-5 \cite{726791} model with 60,100 training images and 10,000 test images, where 60,000 training images and 10,000 test images are from the Fashion-MNIST dataset \cite{abs-1708-07747}, and the other 100 training images are from the MNIST \cite{lecun1998mnist} dataset.
Here, the extra 100 training images are used as watermark key samples.
Similarly, for the CIFAR-10 dataset, we train the VGG-16 \cite{SimonyanZ14a} model with 50,100 training images and 10,000 test images, where 50,000 training images and 10,000 test images are from the CIFAR-10 dataset \cite{krizhevsky2009learning}, and the other 100 training images are from the CIFAR-100 \cite{krizhevsky2009learning} dataset.
We use the Keras \cite{chollet2015keras} framework to train the LeNet-5 and VGG-16 models, where the LeNet-5 architecture is from the work \cite{726791}, and the VGG-16 architecture is from \cite{vgg16model2018}.

\textbf{Parameter settings}.
Table \ref{table2} summarizes the parameter settings of generating the watermark key samples.
For the LeNet-5 model \cite{726791}, we selected 100 samples from the MNIST dataset \cite{lecun1998mnist} (class ``0'') as the watermark key samples. For the VGG-16 model \cite{SimonyanZ14a}, we selected 100 samples from the CIFAR-100 dataset \cite{krizhevsky2009learning} (class ``apple'') as the watermark key samples. In addition, 100 user fingerprints ranging from ``user fp0'' to ``user fp99'' are embedded in the 100 watermark key samples respectively.
In the training phase, we train the LeNet-5 model for 50 epochs with the Adam \cite{KingmaB14} optimizer and train the VGG-16 model for 100 epochs with the SGD optimizer.
The initial learning rate and the momentum of the SGD optimizer are set to be 0.1 and 0.9, respectively.
For the two models, we set the batch size to be 128, and select the cross-entropy loss as the loss function.

\begin{table}[!htbp]
\renewcommand{\arraystretch}{1.3}
\caption{Parameter settings of generating the watermark key samples}
\label{table2}
\centering
\begin{tabular}{|c|c|c|}
\hline
Parameters & LeNet-5 & VGG-16 \\
\hline
\tabincell{c}{Image source} & \tabincell{c}{Class ``0'' \\(from MNIST)} & \tabincell{c}{Class ``apple'' \\(from CIFAR-100)}\\
\hline
\tabincell{c}{\# watermark key samples} & 100 & 100\\
\hline
Users' fingerprints & \tabincell{c}{``user fp0$\sim$99''} & \tabincell{c}{``user fp0$\sim$99''}\\
\hline
\end{tabular}
\end{table}

\textbf{Metrics}.
To evaluate the performance of the proposed watermarking method, we adopt the metrics as follows:
\begin{itemize}
\item[-] \textbf{Test accuracy}:
accuracy on the test set, which does not contain the watermark key samples.
\item[-] \textbf{Watermark accuracy (WM accuracy)}:
we use the watermark accuracy \cite{ZhangGJWSHM18} to evaluate the effectiveness of the proposed method on ownership verification.
The WM accuracy denotes DNN's accuracy on the set of watermark key pairs $D_{wm}$.
In our settings, $D_{wm}$ contains 100 watermark key samples.
\item[-] \textbf{Average mean squared error}:
the mean squared error (MSE) \cite{WangBSS04} is a widely used metric for evaluating the quality difference between two images.
It is calculated by the squared difference between pixel values of the original image and pixel values of the modified image.
We use the average MSE of 100 images to evaluate the impact of embedded user's fingerprint on image quality.
\item[-] \textbf{Fingerprint authentication success rate (FASR)}:
we define the fingerprint authentication success rate (FASR) to evaluate the effectiveness of the proposed method on user's fingerprint authentication. Let $N_f$ be the number of users' fingerprints successfully authenticated, and $N$ be the total number of users' fingerprints. The FASR is calculated by $N_f /N$.  In the experiment, the value of $N$ is 100.
\end{itemize}

\subsection{Effectiveness}
We evaluate the effectiveness of the proposed method from two aspects: ownership verification and user's fingerprint authentication.

\textbf{Ownership verification}.
We test the WM accuracy and test accuracy of the LeNet-5 and VGG-16 on Fashion-MNIST and CIFAR-10 datasets, respectively. The experimental result is shown in Table \ref{table3}.
It is shown that, the WM accuracy of the watermarked LeNet-5 and the watermarked VGG-16 are both 100\%, which indicates that the proposed watermarking method can protect DNN's copyright effectively.
Besides, the proposed method will not affect the test accuracy of LeNet-5 and VGG-16 after the watermark is embedded.
For the LeNet-5, the test accuracy before and after embedding the watermark is 90.31\% and 90.35\% respectively.
For the VGG-16, the test accuracy before and after embedding the watermark is 89.07\% and 89.22\%, respectively.
In summary, our watermarking method enables the model owner to verify his or her ownership successfully, while not affecting the performance of the DNN model.

\begin{table}[!htbp]
  \centering
  \renewcommand{\arraystretch}{1.3}
  \caption{Test accuracy and WM accuracy of the watermarked DNN models}
  \resizebox{\columnwidth}{!}{
    \begin{tabular}{|c|c|c|c|}
    \hline
    Dataset & Model & Test accuracy & WM accuracy\\
    \hline
    \multicolumn{1}{|c|}{\multirow{2}*{\tabincell{c}{Fashion-\\MNIST}}} & Clean LeNet-5 & 90.31\% & N/A\\
    \cline{2-4}
      & Watermarked LeNet-5 & 90.35\% & 100\%\\
    \hline
    \multirow{2}[0]{*}{CIFAR-10} & Clean VGG-16 & 89.07\% & N/A \\
    \cline{2-4}
      & Watermarked VGG-16 & 89.22\% & 100\%\\
    \hline
    \end{tabular}}
  \label{table3}
\end{table}

\textbf{User's fingerprint authentication}.
For the user's fingerprint authentication, Table \ref{table4} shows the FASR and average MSE on the LeNet-5 and VGG-16 models, respectively.
The FASR of 100 users' fingerprints on the two DNN models are both 100\%.
The average MSE of 100 users' fingerprints are 0.044 (LeNet-5 model) and 0.015 (VGG-16 model), respectively, which indicates that the image steganography method (i.e., LSB) will not degrade the image quality.
Therefore, the proposed method can support the user's fingerprint authentication without affecting the quality of the image.

\begin{table}[!htbp]
  \centering
  \renewcommand{\arraystretch}{1.3}
  \caption{FASR and average MSE of 100 users' fingerprints}
    \begin{tabular}{|c|c|c|}
    \hline
    Model & FASR & Average MSE \\
    \hline
    LeNet-5 & 100\% & 0.044 \\
    \hline
    VGG-16 & 100\% & 0.015 \\
    \hline
    \end{tabular}
  \label{table4}
\end{table}

\subsection{Robustness}
In this section, we evaluate the robustness of the proposed method against the model fine-tuning attack \cite{Pittaras2017}, the model pruning attack \cite{HanPTD15}, and the query modification attack \cite{NambaS19}.

\textbf{Model fine-tuning attack \cite{Pittaras2017}}.
Model fine-tuning is a common strategy to modify the model by using a small amount of training data, while maintaining the classification performance of the model.
The adversary may try to destroy the watermark in the DNN model through model fine-tuning.
We selected 50\% of data from the test set of the original dataset (Fashion-MNIST and CIFAR-10, respectively) as the fine-tuning training data, and the remaining 50\% of data is used as the test data.
Each watermarked model is fine-tuned for 30 epochs with the SGD optimizer.
The robustness of the proposed method against the model fine-tuning attack is shown in Table \ref{table5}.
It is shown that the WM accuracy of the fine-tuned model is still 100\%.
There are a large number of redundant parameters in DNN, which makes the DNN model robust to model fine-tuning attacks \cite{ZhangGJWSHM18}.
In addition, since our watermark is embedded into the DNN as an extra class, model fine-tuning cannot remove this class.
Therefore, the proposed watermarking method is robust to the fine-tuning attack.

\begin{table}[htbp]
  \centering
  \renewcommand{\arraystretch}{1.3}
  \caption{Robustness against the model fine-tuning attack}
  \resizebox{\columnwidth}{!}{
    \begin{tabular}{|c|c|c|c|}
    \hline
    Dataset & Model & Test accuracy & \tabincell{c}{WM accuracy} \\
    \hline
    \multicolumn{1}{|c|}{\multirow{2}*{\tabincell{c}{Fashion-\\MNIST}}} & \tabincell{c}{Watermarked LeNet-5}  & 90.35\% & 100\% \\
    \cline{2-4}
      & \tabincell{c}{Fine-tuned LeNet-5}  & 90.34\% & 100\% \\
    \hline
    \multicolumn{1}{|c|}{\multirow{2}*{CIFAR-10}} & \tabincell{c}{Watermarked VGG-16 } & 89.22\% & 100\% \\
    \cline{2-4}
      & \tabincell{c}{Fine-tuned VGG-16} & 89.36\% & 100\% \\
    \hline
    \end{tabular}}
  \label{table5}
\end{table}

\begin{figure}[!t]
\centering
\subfloat[Fashion-MNIST]{\includegraphics[width=2.4in]{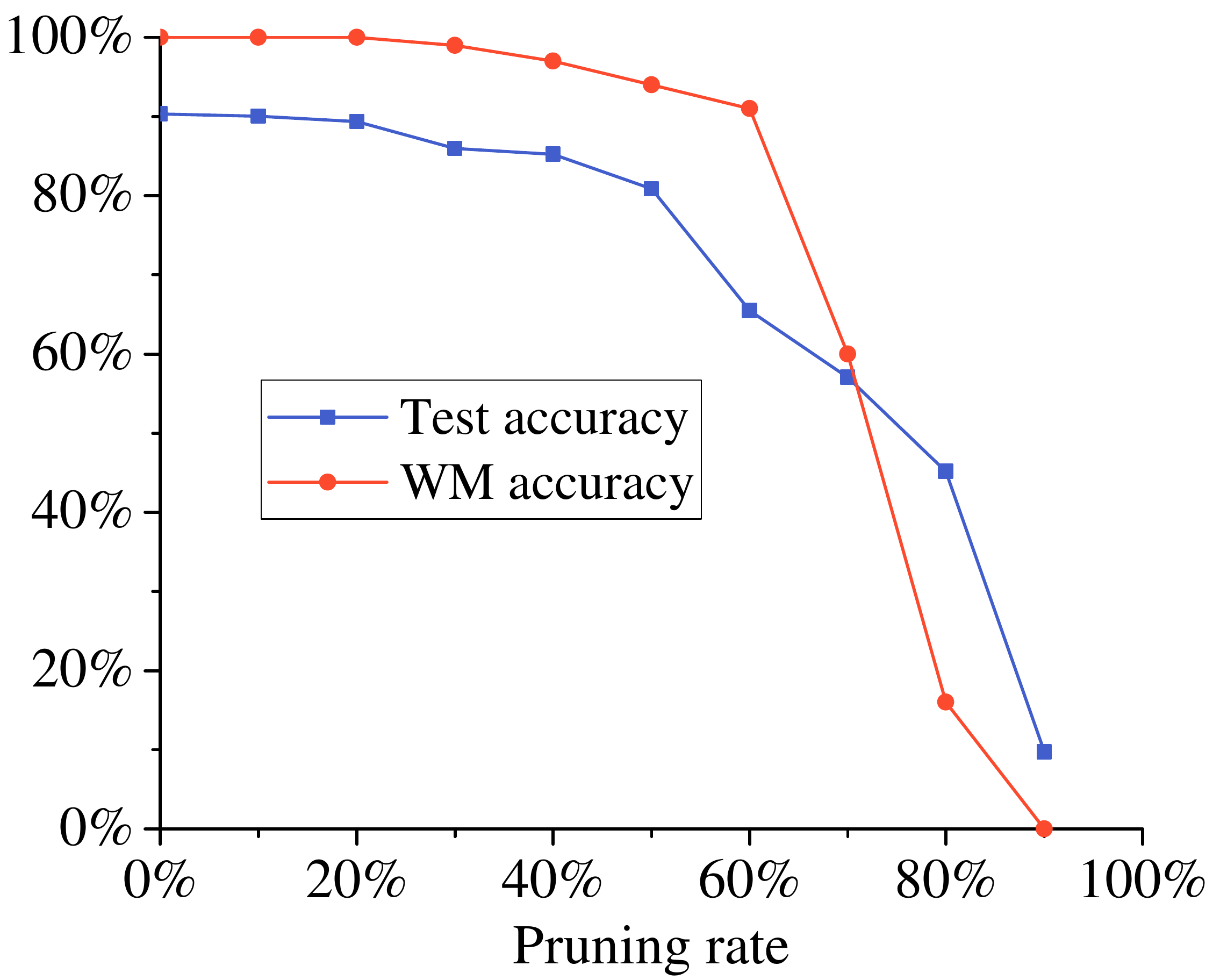}
\label{fig3-1}}
\hfil
\subfloat[CIFAR-10]{\includegraphics[width=2.4in]{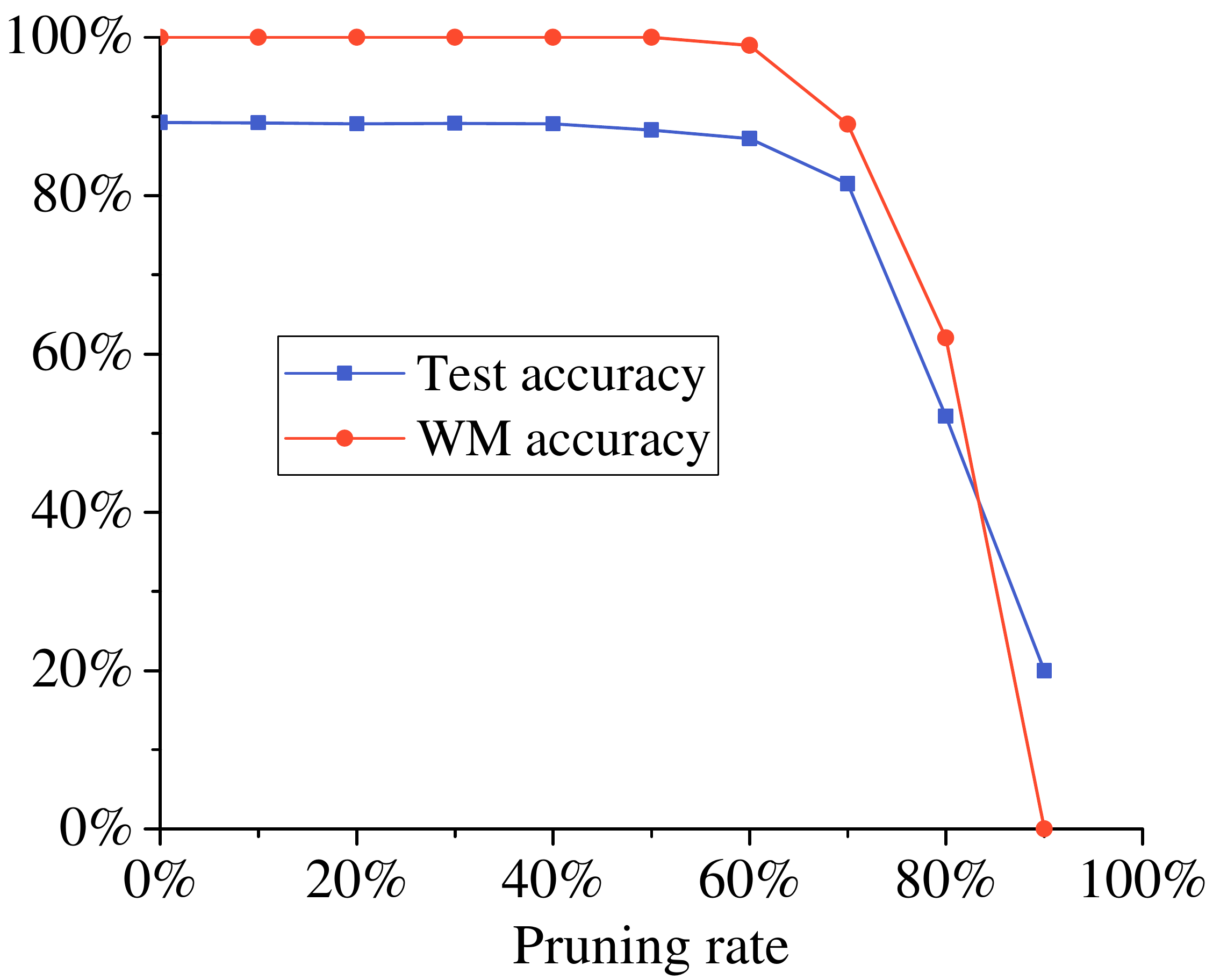}
\label{fig3-2}}
\caption{Robustness against the model pruning attack.}
\label{fig3}
\end{figure}

\textbf{Model pruning attack \cite{HanPTD15}}.
Model pruning is a conventional method of compressing deep neural networks. The adversary may try to remove the potential watermark in a deep neural network by compressing the redundant neurons.
We adopt the the pruning method in \cite{HanPTD15} to prune all convolutional layers and fully-connected layers of the watermarked DNN.
Specifically, we first sort the weights of each layer by the absolute value, and then set $z$\% weights that have the lowest absolute values to 0, where $z$\% denotes the pruning rate.
The robustness of the proposed watermarking method against the model pruning attack is shown in Fig. \ref{fig3}.
The watermark will not be removed if only a few parameters are pruned.
For instance, when the pruning rate $z\%$ is less than 60\%, the WM accuracy of LeNet-5 model is greater than 91\%, and the WM accuracy of VGG-16 model is greater than 99\%.
As $z\%$ increases, the WM accuracy of the pruned model will decrease.
However, the attacker will not prune more than 60\% of the model parameters. The reason is that, the model's test accuracy will drop sharply when the pruning rate exceeds 60\%.
In conclusion, the proposed watermarking method is robust to the model pruning attack.

\textbf{Query modification attack \cite{NambaS19}}.
Query modification attack is a novel attack method where the adversary employs an autoencoder to detect and remove the watermarks in watermark key samples.
Such an attack can invalidate the watermark query during ownership verification.
We adopt an autoencoder from \cite{autoencoder2018} to attack the watermark key samples.
The robustness of the proposed method against the query modification attack is shown in Table \ref{table6}.
The WM accuracy on the attacked set of watermark key pairs is 100\% (LeNet-5) and 85\% (VGG-16), which indicates that our method is robust to the query modification attack.
Note that, the WM accuracy of the VGG-16 model drops slightly. The reason is as follows.
We only use 10\% of the training data to train the autoencoder. The images in the Fashion-MNIST dataset are grayscale images, and the trained autoencoder can obtain excellent generalization capability.
However, the images in the CIFAR-10 dataset are colored images, and the trained autoencoder performs slightly worse. In other words, on the CIFAR-10 dataset, the autoencoder cannot accurately restore each image while removing noises.
For some watermarked key images, the autoencoder cannot restore them accurately, which makes the WM accuracy of the VGG-16 model on the watermarked key images slightly decrease after the query modification attack.

\begin{table}[htbp]
  \centering
  \renewcommand{\arraystretch}{1.3}
  \caption{Robustness against the query modification attack.}
    \begin{tabular}{|c|c|c|}
    \hline
    \multirow{2}*{Model} & \multicolumn{2}{c|}{WM accuracy}  \\
    \cline{2-3}
                         & Before attack & After attack \\
    \hline
    LeNet-5 & 100\% & 100\% \\
    \hline
    VGG-16 & 100\% & 85\% \\
    \hline
    \end{tabular}
  \label{table6}
\end{table}

\subsection{Comparison with Existing Works}
In this section, we compare the proposed method with the existing works, including the logo-based watermarking method \cite{ZhangGJWSHM18}, the noise-based watermarking method \cite{ZhangGJWSHM18}, and the adversarial frontier stitching (AFS) watermarking method \cite{MerrerPT20}.
We use the project \cite{wmR2020} to implement the logo-based and the noise-based watermarking methods locally, and use the project \cite{afs2021} to implement the AFS watermarking method.
The watermarks are embedded into the DNN models under the same training settings.

\begin{figure}[!t]
\centering
\subfloat[LeNet-5]{\includegraphics[width=2.3in]{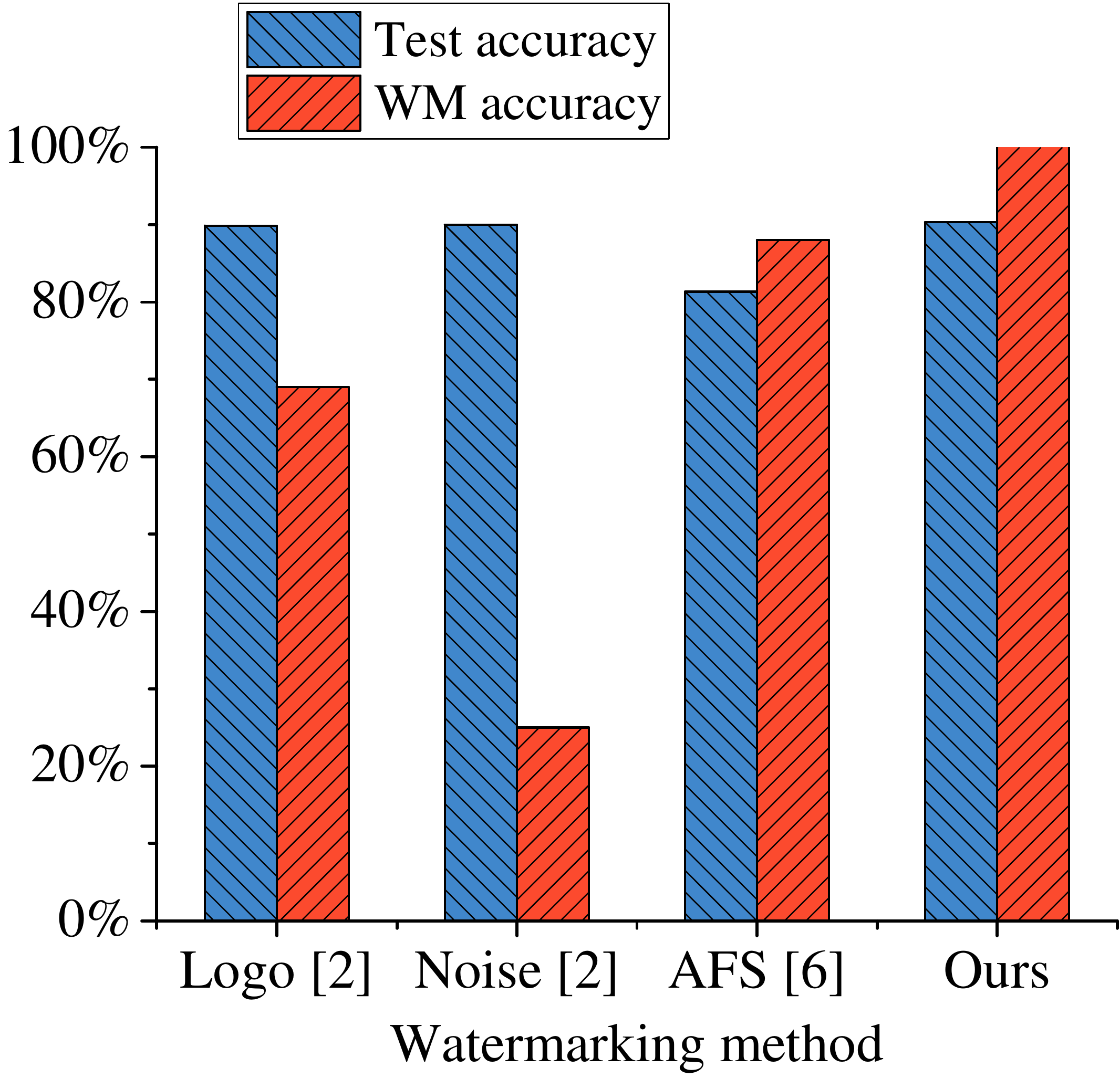}
\label{fig5-1}}
\hfil
\subfloat[VGG-16]{\includegraphics[width=2.3in]{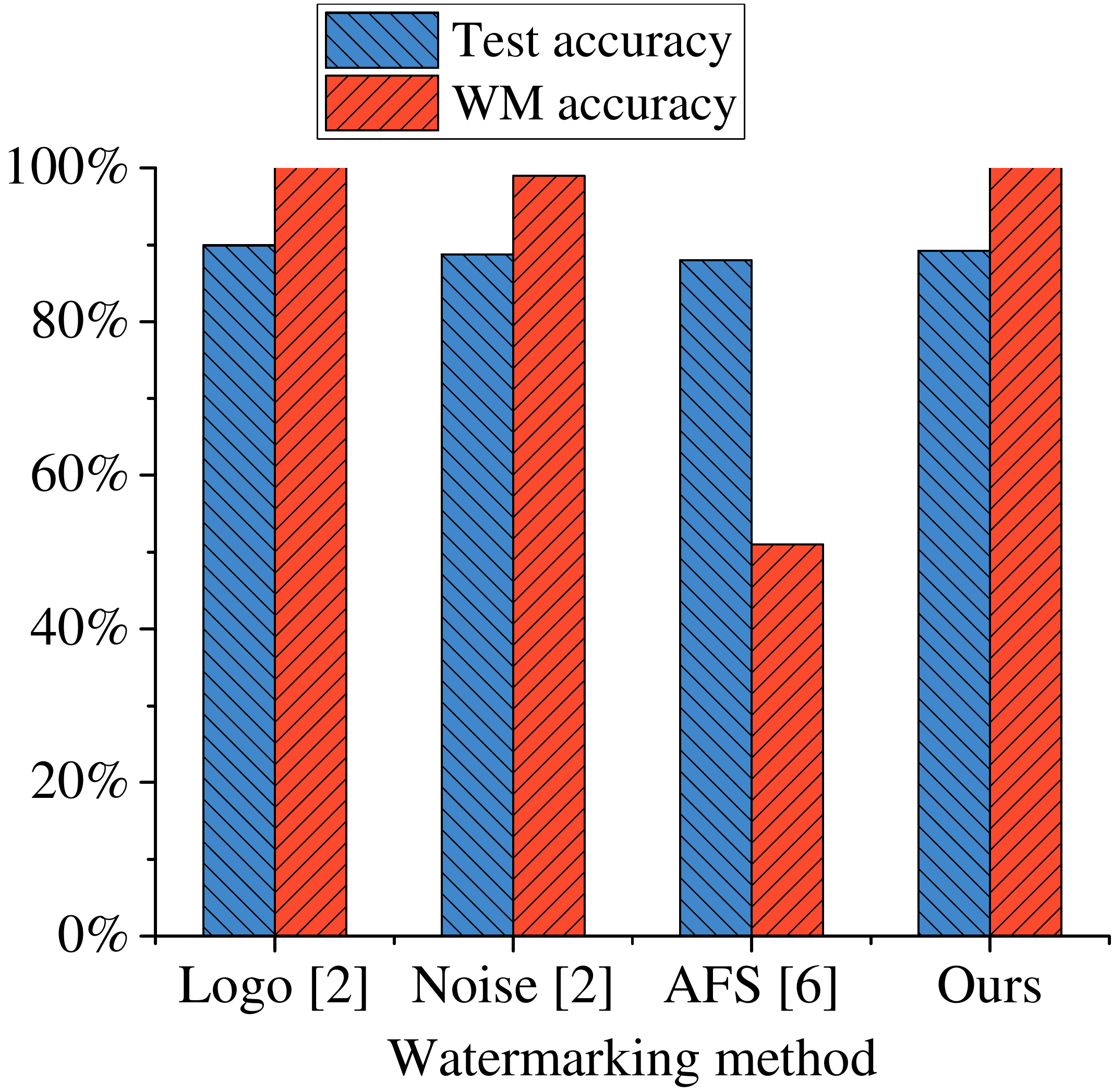}
\label{fig5-2}}
\caption{Test accuracy and WM accuracy comparison for different watermarking methods.}
\label{fig5}
\end{figure}

\textbf{Performance comparison}.
We compare our method with the existing three watermarking methods in terms of test accuracy and WM accuracy, as shown in Fig. \ref{fig5}.
Our watermarking method has the highest test accuracy, which is the closest to the test accuracy of the clean model.
For the WM accuracy, as shown in Fig. \ref{fig5}(a), the WM accuracy of the logo-based, noise-based, and AFS watermarking methods on the LeNet-5 model is 69\%, 25\%, and 88\%, respectively, while our watermarking method obtains 100\% WM accuracy on the LeNet-5 model.
In Fig. \ref{fig5}(b), the WM accuracy of the logo-based, noise-based, and AFS watermarking methods on the VGG-16 model is 100\%, 99\%, and 51\%, respectively, and the WM accuracy of our watermarking method on the VGG-16 model is 100\%.
In summary, our method performs better than existing watermarking methods \cite{ZhangGJWSHM18, MerrerPT20} on test accuracy and WM accuracy.

\begin{figure}[!t]
\centering
\includegraphics[width=3.4in]{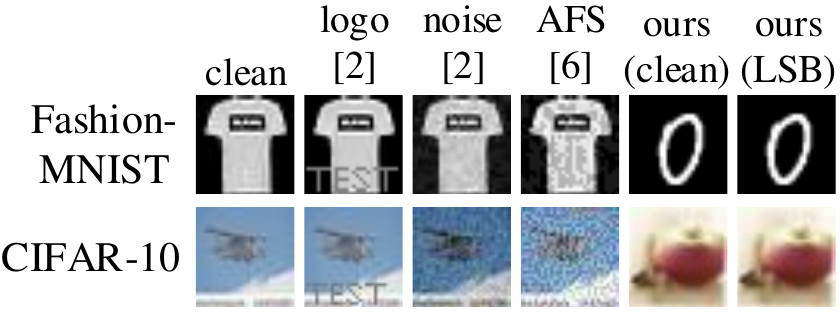}
\caption{An example of the watermark key sample generated by different watermarking methods.}
\label{fig4}
\end{figure}

\textbf{Concealment comparison of watermark key samples}.
Fig. \ref{fig4} shows an example of the watermark key sample generated by the four watermarking methods.
It can be seen that, the watermark key samples generated by the logo-based, the noise-based, and the AFS watermarking methods degrade the image quality obviously.
Thus, it is easy for an attacker to perceive the existence of the watermark.

For the four watermarking methods, we calculate the MSE values between the watermark key samples and the original images.
The average MSE of 100 watermark key samples generated by the four watermarking methods is shown in Table \ref{table7}.
It is shown that, our watermarking method will not compromise the image quality, as the average MSE value is extremely small.
The average MSE value between the steganographic image and the original image is 0.044 (Fashion-MNIST) and 0.015 (CIFAR-10), respectively.
However, the MSE values of the existing watermarking methods \cite{ZhangGJWSHM18, MerrerPT20} are high, which indicates that the watermark patterns embedded in watermark key images can be easily noticed by an attacker.
Therefore, compared with the existing watermarking methods \cite{ZhangGJWSHM18, MerrerPT20}, the watermark key samples generated by the proposed method are more concealed.

\begin{table}[!htbp]
  \centering
  \renewcommand{\arraystretch}{1.3}
  \caption{Average MSE comparison for different types of watermarking methods}
    \begin{tabular}{|c|c|c|}
    \hline
    Dataset & Watermark type  & Average MSE \\
    \hline
    \multicolumn{1}{|c|}{\multirow{4}[0]{*}{Fashion-MNIST}} & Logo \cite{ZhangGJWSHM18} & 718.27 \\
    \cline{2-3}
      & Noise \cite{ZhangGJWSHM18} & 291.7 \\
    \cline{2-3}
      & AFS \cite{MerrerPT20} & 454.82 \\
    \cline{2-3}
      & \textbf{Ours} & \textbf{0.044} \\
    \hline
    \multirow{4}[0]{*}{CIFAR-10} & Logo \cite{ZhangGJWSHM18} & 203.58 \\
    \cline{2-3}
      & Noise \cite{ZhangGJWSHM18} & 298.96 \\
    \cline{2-3}
      & AFS \cite{MerrerPT20} & 629.92 \\
    \cline{2-3}
      & \textbf{Ours} & \textbf{0.015} \\
    \hline
    \end{tabular}
  \label{table7}
\end{table}

\begin{figure}[!t]
\centering
\subfloat[Fashion-MNIST]{\includegraphics[width=3.1in]{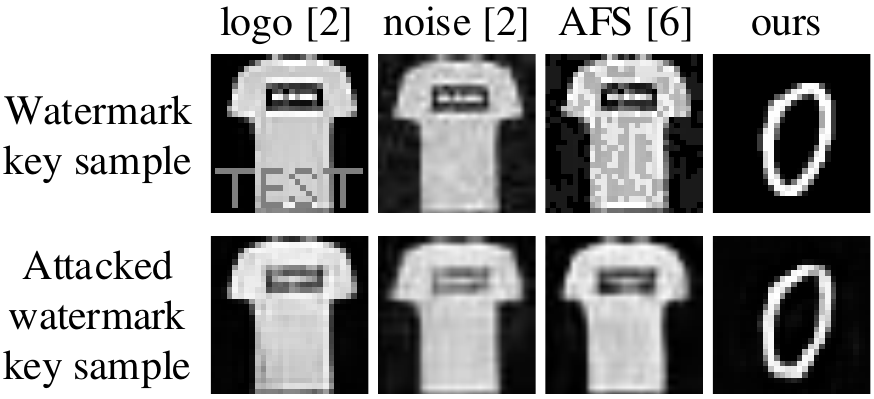}}
\hfil
\subfloat[CIFAR-10]{\includegraphics[width=3.1in]{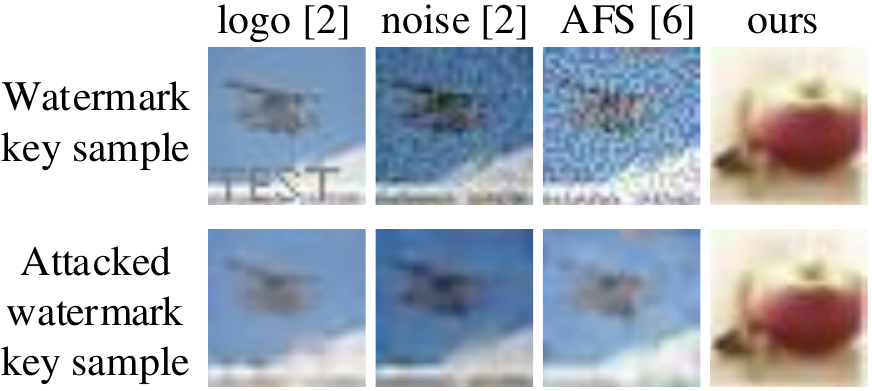}}
\caption{Robustness comparison of four watermarking methods against the query modification attack.}
\label{fig6}
\end{figure}

\textbf{Robustness against the query modification attack}.
We compare the robustness of four watermarking methods against the query modification attack.
Fig. \ref{fig6} shows an example image reconstructed by the autoencoder for different types of watermarking methods.
It is shown that, the ``test'' logo, the noise, and the adversarial perturbation added in the image are removed by the autoencoder.
However, the query modification attack is invalid for our watermarking method.
The reason is as follows.
Our method uses an extra class to watermark the DNN model.
The watermark key samples are some clean images outside the original training set, rather than particular patterns superimposed on the clean images, thus they can resist the query modification attacks.
In contrast, for the existing methods \cite{ZhangGJWSHM18, MerrerPT20}, the watermark key samples are generated by superimposing some particular patterns (e.g., logo patterns and noise patterns) on the images, thus the attacker can invalidate these watermark keys by modifying the watermark key samples using autoencoder.

Table \ref{table8} shows the WM accuracy of these four watermarking methods under the query modification attack.
The logo-based and the noise-based watermarking methods are not robust to the query modification attack, as the WMK accuracy of these two methods is lower than 16\% after the attack.
The AFS watermarking method performs well (91\% WM accuracy) on the LeNet-5 model, but it only obtains a WM accuracy of 35\% on the VGG-16 model.
Our watermarking method can obtain 100\% (LeNet-5) and 85\% (VGG-16) WM accuracy, which is significantly higher than the WM accuracy of the other three watermarking methods.
In conclusion, our method is robust to the query modification attacks.

\begin{table}[htbp]
  \centering
  \renewcommand{\arraystretch}{1.3}
  \caption{WM accuracy comparison for different watermarking methods against the query modification attack}
    \begin{tabular}{|c|c|c|c|}
    \hline
    \multirow{2}*{Model} & \multirow{2}*{\tabincell{c}{Watermarking \\ method}} & \multicolumn{2}{c|}{WM accuracy} \\
    \cline{3-4}
    & & Before attack & After attack \\
    \hline
    \multicolumn{1}{|c|}{\multirow{4}[0]{*}{LeNet-5}} & Logo \cite{ZhangGJWSHM18} & 69\% & 11\% \\
    \cline{2-4}
      & Noise \cite{ZhangGJWSHM18} & 25\% & 14\% \\
    \cline{2-4}
      & AFS \cite{MerrerPT20} & 88\% & 91\% \\
    \cline{2-4}
      & \textbf{Ours} & \textbf{100\%} & \textbf{100\%} \\
    \hline
    \multirow{4}[0]{*}{VGG-16} & Logo \cite{ZhangGJWSHM18} & 100\% & 15\% \\
    \cline{2-4}
      & Noise \cite{ZhangGJWSHM18} & 99\% & 16\% \\
    \cline{2-4}
      & AFS \cite{MerrerPT20} & 51\% & 35\% \\
    \cline{2-4}
      & \textbf{Ours} & \textbf{100\%} & \textbf{85\%} \\
    \hline
    \end{tabular}
  \label{table8}
\end{table}

\section{Conclusion}
We propose a method of using an additional class and steganographic images to embed the watermark into the DNN model, which enables our method to support both ownership verification and user's fingerprint authentication.
Experimental evaluations on Fashion-MNIST and CIFAR-10 datasets demonstrate that the proposed watermarking method can achieve 100\% WM accuracy and 100\% fingerprint authentication success rate without affecting the test accuracy of the model.
Meanwhile, the proposed watermarking method is robust to the model fine-tuning attack, model pruning attack, and query modification attack.
Compared with the existing logo-based \cite{ZhangGJWSHM18}, noise-based \cite{ZhangGJWSHM18} and AFS \cite{MerrerPT20} watermarking methods, the proposed watermarking method has better performance on test accuracy and WM accuracy, and is more robust to the query modification attack.
In future work, we will explore the robustness of the proposed method against the fingerprint collusion attack.

\bibliographystyle{IEEEtran}
\bibliography{ref}

\begin{thebibliography}{10}
\providecommand{\url}[1]{#1}
\csname url@samestyle\endcsname
\providecommand{\newblock}{\relax}
\providecommand{\bibinfo}[2]{#2}
\providecommand{\BIBentrySTDinterwordspacing}{\spaceskip=0pt\relax}
\providecommand{\BIBentryALTinterwordstretchfactor}{4}
\providecommand{\BIBentryALTinterwordspacing}{\spaceskip=\fontdimen2\font plus
\BIBentryALTinterwordstretchfactor\fontdimen3\font minus
  \fontdimen4\font\relax}
\providecommand{\BIBforeignlanguage}[2]{{%
\expandafter\ifx\csname l@#1\endcsname\relax
\typeout{** WARNING: IEEEtran.bst: No hyphenation pattern has been}%
\typeout{** loaded for the language `#1'. Using the pattern for}%
\typeout{** the default language instead.}%
\else
\language=\csname l@#1\endcsname
\fi
#2}}
\providecommand{\BIBdecl}{\relax}
\BIBdecl

\bibitem{RibeiroGC15}
M.~Ribeiro, K.~Grolinger, and M.~A.~M. Capretz, ``{MLaaS}: {M}achine learning
  as a service,'' in \emph{Proceedings of the 14th {IEEE} International
  Conference on Machine Learning and Applications}, 2015, pp. 896--902.

\bibitem{ZhangGJWSHM18}
J.~Zhang, Z.~Gu, J.~Jang, H.~Wu, M.~P. Stoecklin, H.~Huang, and I.~Molloy,
  ``Protecting intellectual property of deep neural networks with
  watermarking,'' in \emph{Proceedings of the Asia Conference on Computer and
  Communications Security}, 2018, pp. 159--172.

\bibitem{Pittaras2017}
N.~Pittaras, F.~Markatopoulou, V.~Mezaris, and I.~Patras, ``Comparison of
  fine-tuning and extension strategies for deep convolutional neural
  networks,'' in \emph{Proceedings of the 23rd International Conference on
  MultiMedia Modeling}, 2017, pp. 102--114.

\bibitem{HanPTD15}
S.~Han, J.~Pool, J.~Tran, and W.~J. Dally, ``Learning both weights and
  connections for efficient neural network,'' in \emph{Proceedings of the
  Advances in Neural Information Processing Systems}, 2015, pp. 1135--1143.

\bibitem{UchidaNSS17}
Y.~Uchida, Y.~Nagai, S.~Sakazawa, and S.~Satoh, ``Embedding watermarks into
  deep neural networks,'' in \emph{Proceedings of the {ACM} on International
  Conference on Multimedia Retrieval}, 2017, pp. 269--277.

\bibitem{MerrerPT20}
E.~L. Merrer, P.~P{\'{e}}rez, and G.~Tr{\'{e}}dan, ``Adversarial frontier
  stitching for remote neural network watermarking,'' \emph{Neural Computing
  and Applications}, vol.~32, no.~13, pp. 9233--9244, 2020.

\bibitem{AdiBCPK18}
Y.~Adi, C.~Baum, M.~Ciss{\'{e}}, B.~Pinkas, and J.~Keshet, ``Turning your
  weakness into a strength: {W}atermarking deep neural networks by
  backdooring,'' in \emph{Proceedings of the 27th {USENIX} Security Symposium},
  2018, pp. 1615--1631.

\bibitem{NambaS19}
R.~Namba and J.~Sakuma, ``Robust watermarking of neural network with
  exponential weighting,'' in \emph{Proceedings of the {ACM} Asia Conference on
  Computer and Communications Security}, 2019, pp. 228--240.

\bibitem{lsbsteganography2020}
R.~David, ``{LSB-S}teganography,''
  \url{https://github.com/RobinDavid/LSB-Steganography}, 2020.

\bibitem{abs-1708-07747}
H.~Xiao, K.~Rasul, and R.~Vollgraf, ``Fashion-{MNIST}: a novel image dataset
  for benchmarking machine learning algorithms,'' \emph{arXiv:1708.07747},
  2017.

\bibitem{krizhevsky2009learning}
A.~Krizhevsky, G.~Hinton \emph{et~al.}, ``Learning multiple layers of features
  from tiny images,'' \emph{Technical Report}, 2009.

\bibitem{WangBSS04}
Z.~Wang, A.~C. Bovik, H.~R. Sheikh, and E.~P. Simoncelli, ``Image quality
  assessment: {F}rom error visibility to structural similarity,'' \emph{{IEEE}
  Transactions on Image Processing}, vol.~13, no.~4, pp. 600--612, 2004.

\bibitem{726791}
Y.~{Lecun}, L.~{Bottou}, Y.~{Bengio}, and P.~{Haffner}, ``Gradient-based
  learning applied to document recognition,'' \emph{Proceedings of the {IEEE}},
  vol.~86, no.~11, pp. 2278--2324, 1998.

\bibitem{SimonyanZ14a}
K.~Simonyan and A.~Zisserman, ``Very deep convolutional networks for
  large-scale image recognition,'' in \emph{Proceedings of the 3rd
  International Conference on Learning Representations}, 2015, pp. 1--14.

\bibitem{RouhaniCK19}
B.~D. Rouhani, H.~Chen, and F.~Koushanfar, ``Deep{S}igns: {A}n end-to-end
  watermarking framework for ownership protection of deep neural networks,'' in
  \emph{Proceedings of the 24th International Conference on Architectural
  Support for Programming Languages and Operating Systems}, 2019, pp. 485--497.

\bibitem{ZhongZ0G020}
Q.~Zhong, L.~Y. Zhang, J.~Zhang, L.~Gao, and Y.~Xiang, ``Protecting {IP} of
  deep neural networks with watermarking: {A} new label helps,'' in
  \emph{Proceedings of the Advances in Knowledge Discovery and Data Mining -
  24th Pacific-Asia Conference}, 2020, pp. 462--474.

\bibitem{lecun1998mnist}
Y.~LeCun, C.~Cortes, and C.~J. Burges, ``The {MNIST} database of handwritten
  digits,'' \emph{http://yann. lecun. com/exdb/mnist/}, 1998.

\bibitem{chollet2015keras}
F.~Chollet \emph{et~al.}, ``Keras,'' \url{https://keras.io}, 2015.

\bibitem{vgg16model2018}
W.~Jiang, ``Keras-cifar10,'' \url{https://github.com/jerett/Keras-CIFAR10},
  2018.

\bibitem{KingmaB14}
D.~P. Kingma and J.~L. Ba, ``Adam: {A} method for stochastic optimization,'' in
  \emph{Proceedings of the 3rd International Conference on Learning
  Representations}, 2015, pp. 1--15.

\bibitem{autoencoder2018}
``{DeepLearningDenoise},''
  \url{https://github.com/shibuiwilliam/DeepLearningDenoise}, 2018.

\bibitem{wmR2020}
``Watermark{R}obustness,''
  \url{https://github.com/CodeSubmission642/WatermarkRobustness}, 2020.

\bibitem{afs2021}
T.~von K{\"a}nel, ``adversarial-frontier-stitching,''
  \url{https://github.com/dunky11/adversarial-frontier-stitching}, 2021.

\end{thebibliography}

\end{document}